# RL-Ballast: Ship Ballast Water Path Planning and Clog Prediction via Reinforcement Learning

Ming-Kuan Lin, Yi-Chung Lai, Ming-Hsin Chiang, Tsung-Wei Pan, Jung-Hua Wang*

***Abstract***—Under the Shipping 4.0 paradigm, autonomous and reduced-crew vessels require intelligent internal systems to ensure operational safety and structural stability. Ballast-water control is essential for maintaining ship trim and integrity, yet conventional rule-based logic and manual operations have limited adaptability to unexpected hydraulic anomalies, such as valve failures or pipe blockages, and often require dense pressure or flow sensors for diagnosis. To address these limitations, this paper proposes RL-Ballast, a graph-based deep reinforcement learning framework for adaptive ballast-water path planning and sensor-frugal blockage candidate scoring. The framework converts the valve-permutation problem into 54 feasible fluid-transfer routes generated by graph theory and depth-first search. It approximates the partially observable ballast environment using frame-stacked tank levels and action outcomes, enabling the agent to infer hidden blockage effects without explicitly modeling a high-dimensional POMDP. Episode-level failed-action memory and dynamic action masking are integrated during deterministic inference to prevent repeated ineffective actions and support immediate rerouting. Failed transfer histories are accumulated to rank suspicious valves or pipe segments without dense flow or pressure instrumentation. Monte Carlo evaluations in a simulated ballast system show that RL-Ballast completes all unexpected single-blockage scenarios. Compared with a Dijkstra rule-based baseline, it reduces the average decision steps from 61.0 to 41.5. For diagnostic support, the failure-history scoring scheme achieves a 100% Top-3$_{(hit)}$ rate, a 66.7% strict Top-1$_{(hit)}$ rate, and an 83.3% Top-1$_{(tie\text{-}hit)}$ rate under serially indistinguishable blockage conditions. These results indicate that RL-Ballast provides adaptive rerouting and maintenance-oriented blockage diagnosis under limited sensing conditions.

Submitted on June. 30, 2026 for review, this work was supported in part by the Ship and Ocean Industries R&D Center (SOIC). *(Corresponding author: Jung-Hua Wang).*

*Jung-Hua Wang* is with the AI Research Center, National Taiwan Ocean University, Keelung 20224, Taiwan (jhwang@email.ntou.edu.tw).

*Ming-Hsin Chiang* was with Department of Electrical Engineering, National Taiwan Ocean University, Keelung 20224, Taiwan. He is now with Shinsoft Corporation, Taiwan (ac872xy@gmail.com).

*Yi-Chung Lai* was with Department of Electrical Engineering, National Taiwan Ocean University, Keelung 20224, Taiwan. He is now with AI Research Center (e-mail: k8130488@gmail.com).

*Ming-Kuan Lin* is currently with Department of Electrical Engineering, National Taiwan Ocean University, Keelung 20224, Taiwan (e-mail: jacksonlin092@gmail.com).

## I. INTRODUCTION

Under the Shipping 4.0 paradigm, the maritime industry is moving toward digitalized, intelligent, and increasingly autonomous vessel operations (Sullivan et al., 2021, 2020). Considerable progress has been made in autonomous navigation, collision avoidance, trajectory tracking, and route optimization. However, practical ship autonomy requires not only external navigational intelligence but also reliable control of internal onboard systems. These internal systems are critical for maintaining vessel safety, operational continuity, and reduced-crew ship management.

Among onboard internal systems, the ballast-water piping system is a critical network of pipes, valves, and pumps on a vessel used to load, transfer, and discharge seawater. It ensures structural integrity, controls trim and stability, and integrates with a Ballast Water Treatment System (BWTS) to prevent the spread of invasive species. During loading, unloading, voyage, and special operations, ballast conditions must be adjusted to maintain a safe vessel attitude and structural balance. Improper ballast operations may result in excessive hull stress, poor trim, reduced maneuverability, or hazardous stability conditions. Therefore, intelligent ballast water control is a fundamental requirement for future autonomous and reduced-crew vessels.

In current maritime practice, ballast operations are predominantly executed via manual control or fixed programmable logic controller (PLC) sequences. While effective under nominal operating conditions, these rule-based approaches presuppose the total functionality of all pumps, valves, and pipelines. This assumption introduces operational vulnerabilities in aging vessels, where biofouling, sediment accumulation, and valve degradation frequently cause unforeseen flow restrictions. When such anomalies occur, rigid operational sequences often repeatedly execute ineffective transfer routes, leaving human operators to infer system faults from delayed or stagnant tank-level dynamics. Compounding this challenge, the sparse sensor instrumentation of most existing vessels, which lack dense pressure or flow instrumentation along individual pipe segments, renders direct fault localization difficult. Consequently, fault diagnosis is restricted to indirect inference from macroscopic observations (e.g., volumetric tank variations), making the process time-consuming and heavily reliant on operator intuition. To mitigate these limitations, there is a critical need for a sensor-frugal framework capable of both adaptively rerouting ballast water under unknown blockage or clogging conditions and localizing clogs from sparse operational data. Recent literature has leveraged machine learning for shipboard fluid-system diagnostics and rupture detection. However, these studies primarily use supervised and semi-supervised learning for detection (Ferreno-Gonzalez et al., 2025; Velasco-Gallego et al., 2024) and rely heavily on continuous time-series data from standard automation systems, such as pump motor current, rotational speed, power consumption, and inlet/outlet pressures. Besides, they focus on isolated classification or detection tasks rather than sequential routing optimization.

Reinforcement learning (RL) (Sutton and Barto, 2018) provides a promising approach for this type of sequential decision-making problem. Unlike supervised learning methods, which usually require labeled fault datasets, an RL agent can learn control policies through interaction with a simulated ballast environment. By selecting different water-transfer paths and observing whether the target tank levels increase, the agent can gradually learn to avoid ineffective routes and search for alternative paths. Moreover, failed transfer attempts provide useful diagnostic evidence because repeated failures involving common valves or pipe segments may indicate potential blockage locations. In recent decades, although RL has been widely investigated in maritime applications such as autonomous surface vessel control, collision avoidance, and trajectory tracking, most existing studies have focused on external navigational tasks. The application of RL to internal shipboard fluid systems remains relatively underexplored. Furthermore, existing fault diagnosis methods for shipboard systems often rely on dense sensing or labeled fault data and usually do not actively reconfigure the control policy after a fault is detected. Few studies have jointly addressed adaptive ballast-water path planning and blockage inference under unknown blockage and limited sensing conditions.

To bridge this gap, this paper introduces **RL-Ballast**, a graph-based RL framework designed for dynamic path planning and blockage candidate ranking within maritime ballast water systems. The framework first models the ballast piping network as a topology-constrained graph, deriving feasible water-transfer paths as discrete actions. The ballast routing challenge is formulated as a partially observable Markov decision process (POMDP), where an RL agent is trained to select optimal valve-operation paths based on tank-level observations and transfer outcomes (Bellman, 1957; Kaelbling et al., 1998). To counter the partial observability introduced by hidden blockages, RL-Ballast integrates episode-level failed-action memory alongside dynamic action masking during inference. Furthermore, the system analyzes historical transfer failures to calculate blockage suspicion scores for the specific valves and pipe segments associated with unsuccessful paths.

The main contributions of this study are summarized as follows: (C1) A topology-constrained graph representing feasible water-transfer paths, given a vessel's ballast piping layout, can be automatically generated to enable discrete reinforcement learning actions. (C2) An RL agent specially tailored for dynamic path planning and blockage candidate ranking is designed to perform adaptive ballast-water transfer under unknown blockage conditions. (C3) An episode-level failed-action memory and dynamic action masking mechanism can avoid repeated selection of ineffective (likely clogged) paths during deterministic inference. (C4) A failure history-based blockage candidate scoring method is designed to infer possible clogged valves or pipe segments from unsuccessful transfer attempts, providing sensor-efficient diagnostic support.

The remainder of this paper is organized as follows. Section II reviews related studies on reinforcement learning for maritime systems, shipboard internal system resilience, and fault diagnosis in fluid systems. Section III formulates the ballast water transfer problem using graph representation and reinforcement learning. Section IV describes how to tailor an RL agent for path planning and blockage candidate ranking, failed-action memory, dynamic action masking, and blockage candidate scoring method. Section V reports the simulation experiments and performance analysis. Section VI discusses the implications, limitations, and future extensions of the proposed method. Finally, Section VII concludes this paper.

## II. RELATED WORK

### *A. Reinforcement Learning for Maritime and Shipboard Systems*

RL has been extensively investigated for maritime autonomous systems, particularly in navigation-related tasks. Existing studies have applied RL to autonomous surface vessel trajectory tracking (Wang et al., 2023), collision avoidance (Li et al., 2025), route planning (Geng et al., 2020), and COLREGs-compliant decision-making (Kufoalor et al., 2019). These problems are naturally formulated as sequential decision-making tasks, where an agent must select control actions in uncertain and dynamic maritime environments. Recent reviews of RL-based collision avoidance for unmanned surface vehicles further indicate that value-based, policy-based, and multi-agent RL methods have become important tools for autonomous maritime decision-making. Despite these advances, most existing maritime RL studies focus on external ship behaviors, such as heading control, trajectory following, obstacle avoidance, or ship-to-ship interaction (Sarhadi et al., 2022). The control targets are usually rudder angle, propulsion, heading, speed, or collision-avoidance maneuvers. In contrast, the present study focuses on an internal shipboard fluid system, where the agent must select feasible ballast-water transfer paths under unknown blockage conditions. Therefore, the problem addressed in this paper differs from conventional maritime navigation problems in terms of state representation, action definition, physical constraints, and fault observability.

As a specialized subfield of RL, DRL uses deep neural networks to approximate value functions, policies, or both, enabling agents to learn from high-dimensional or complex state representations. Beyond external navigation, recent studies have begun to explore the use of DRL for internal shipboard cyber–physical systems. Liu et al. (2026) investigated DRL-based resilient restoration of ship cyber–physical systems under uncertain cyberattack scenarios. Their work models cascading failure propagation between communication networks and physical power topology, and uses a masked reinforcement learning strategy to support constrained restoration and load recovery decisions. At first glance, this line of research is relevant to our problem at hand because it demonstrates the potential of DRL for adaptive decision-making in shipboard internal systems rather than only for navigation. However, the target system and failure mechanisms are different from those considered in this study. Shipboard power restoration focuses on electrical topology reconfiguration, load recovery, and cyberattack-induced cascading failures. In contrast, ballast-water path planning involves hydraulic routing, tank-level responses, valve-operation paths, and blockage-induced no-flow conditions. Hence, additional modeling strategies are required to represent ballast piping topology, detect ineffective transfer paths, and infer possible blockage candidates from limited observations.

### *B. AI and RL for Hydraulic and Ballast-Water Operations*

General RL methods have also been explored in hydraulic and water distribution networks (Hu et al., 2020), which share several structural similarities with ballast piping systems, such as network topology, valve operation, pump scheduling, and flow-related constraints. Recent reviews on RL for water resource management indicate that RL is suitable for optimizing sequential decision-making problems in water systems, and that DRL methods such as Deep Q-Network (DQN) (Mnih et al., 2015) are frequently used because many control actions can be discretized. In water distribution networks, DRL has been investigated for pressure control, leakage management, valve scheduling, and pump operation. These studies show that DRL can support operational decision-making in nonlinear hydraulic systems with large action spaces and uncertainty. However, most water distribution studies are designed for land-based municipal infrastructure, where the objectives are typically pressure regulation, leakage reduction, energy efficiency, or contaminant isolation. Ship ballast water systems differ in that the network is embedded in a vessel, the operational objective is tank-level transfer for stability-related tasks, and blockage may be inferred only indirectly from tank-level stagnation. Therefore, existing hydraulic-network RL methods cannot be directly applied without ship-specific graph modeling, path-based action design, and failed-action-based blockage inference.

Artificial intelligence (AI) has also been applied to ballast-water management systems, particularly for optimizing treatment performance and ensuring environmental compliance. For example, online machine learning algorithms have been proposed to continuously update prediction models using sensor data from ships and ports, with the goal of improving ballast water treatment systems that use filtration and ultraviolet disinfection. These studies are important for biological treatment and regulatory compliance of ballast water discharge. However, they address a different aspect of ballast-water management from the present study. This paper focuses on the operational control of ballast-water transfer inside the ship piping network, particularly dynamic path selection and blockage inference under unknown blockage conditions. Therefore, the proposed method complements existing AI-based BWMS studies by addressing the hydraulic routing and fault-aware control aspect of ballast operations.

### *C. Fault Diagnosis in Shipboard fluid Systems*

Fault detection and diagnosis for shipboard fluid systems has also received increasing attention. For ballast-water systems, graph-based deep learning has been explored for fault diagnosis. For example, multi-feature fusion graph convolution methods (Ai et al., 2024) have been proposed to represent ballast-system fault data using probabilistic and correlation topology graphs, allowing graph neural networks to extract structural fault features. Similar data-driven approaches have also been investigated for other shipboard fluid systems. Ferreño-González et al. (2025) studied pipe rupture detection in a shipboard saltwater firefighting system using pressure sensor data and machine learning/deep learning models to distinguish normal and abnormal operating states and locate pipe faults. Deep learning has also been applied to ship pipeline valve leak diagnosis, where end-to-end feature extraction is used to improve fault identification compared with traditional expert-based diagnosis.

Although these studies demonstrate the effectiveness of data-driven fault diagnosis, they are usually formulated as passive monitoring or classification problems. They often require sufficient labeled fault data (Velasco-Gallego et al., 2024), pressure measurements, or sensor arrays. In many existing vessels, dense deployment of pressure or flow sensors along every pipe segment is impractical. Moreover, detecting a fault does not necessarily provide an adaptive control strategy to bypass the fault. In contrast, the present study integrates control and diagnosis: failed transfer attempts are used not only as negative rewards for reinforcement learning but also as diagnostic evidence for ranking possible blockage candidates.

### *D. Research Gap and Algorithmic Rationale*

RL algorithms are broadly categorized into value-based, policy-based, and actor–critic methods. Value-based methods, such as DQN and Double DQN (DDQN) (Hasselt et al., 2016), learn a state- or action-value function to derive an optimal policy, making them well-suited for discrete action-selection problems. Policy-based methods, such as Proximal Policy Optimization (PPO), directly optimize a parameterized policy, offering stable updates for continuous or stochastic control tasks. Actor–critic methods combine both approaches—using an actor to optimize the policy and a critic to estimate the value function, which reduces variance while retaining direct policy optimization. Representative examples include A3C and Soft Actor-Critic (SAC).

Recent reviews (Terven, 2025) emphasize that no single RL algorithm is universally optimal; selection depends heavily on the problem structure, action space, and deployment environment. Given that shipboard ballast-water route planning is naturally formulated as a finite, discrete action-selection task—where each action corresponds to a graph-generated feasible transfer path, this study adopts a value-based deep RL framework. DQN is selected due to its proven success in discrete control and its foundational roots in temporal difference learning. To address the inherent overestimation bias of standard DQN, this study incorporates DDQN, which decouples action selection from action evaluation. While the DQN family includes various powerful extensions (e.g., Dueling DQN, Prioritized Experience Replay, Distributional DQN, and Rainbow DQN) and has even inspired continuous-action variants like QT-Opt (Kalashnikov et al., 2018), the baseline DQN and DDQN architectures provide a highly interpretable and extensible foundation for topology-constrained routing. This setup effectively solves the immediate graph-based routing problem while leaving room for future extensions, such as continuous valve control or actor–critic methods.

While existing studies provide important foundations, several critical gaps remain: (1) **Scope of Maritime RL:** Prior research predominantly focuses on external navigation, collision avoidance, and motion control. While recent cyber–physical studies leverage DRL for internal system resilience, they target electrical power restoration rather than hydraulic ballast transfer; (2) **Fault Diagnosis vs. Active Control:** Existing fault-diagnosis methods for shipboard fluid systems rely heavily on supervised learning, labeled data, or dense sensor networks, and they rarely reconfigure the control policy after a fault is detected; (3) **Domain Misalignment:** RL-based hydraulic control literature focuses almost exclusively on land-based water distribution networks, which cannot directly accommodate ship-specific ballast objectives or blockage inference. Consequently, the integration of graph-based path generation, adaptive routing, dynamic failed-action masking, and failure history-based blockage scoring under unknown conditions remains insufficiently explored.

To the best of our knowledge, this study introduces the first RL-based framework specifically engineered for shipboard ballast-water transfer routing under unknown blockage and limited sensing conditions. Departing from traditional maritime RL applications, such as trajectory tracking or vessel stability, the proposed **RL-Ballast** framework targets internal piping operations. It uniquely integrates graph-based feasible path generation, value-based adaptive routing, failed-action memory, dynamic action masking, and failure history-based blockage candidate scoring to achieve fault-aware, resilient ballast-water transfer.

## III. System Modeling and Problem Formulation

To apply RL to a physical piping network, we must first abstract the physical infrastructure into a computationally viable mathematical model. To achieve this, we employ graph theory to represent the network's topology and utilize a Markov decision process (MDP) to model the underlying control problem.

### *A. Topology-Constrained Directed Graph*

To computationally tackle the fluid routing problem, a physical ship ballast piping network is treated as a directed topology-constrained graph denoted as $G = (V, E)$. The vertex set $V$ encompasses all functional components and is partitioned into four distinct subsets: $V=V_{tank} \cup V_{valve} \cup V_{pump} \cup V_{node}$, representing ballast tanks, operable valves, water pumps, and passive pipe junctions, respectively. The edge set $E \subseteq V \times V$ represents the physical pipe segments connecting these components.

Unlike passive land-based water distribution networks, shipboard ballast transfer is highly dependent on active pump actuation, which imposes strict directionality on fluid flow. Consequently, the graph is represented by an adjacency matrix $M \in \mathbb{R}^{|v|\times|v|}$, where $m_{ij}=1$ if a hydraulically viable direct connection exists from vertex $i$ to vertex $j$, and $m_{ij}=0$ otherwise. While high-fidelity computational fluid dynamics (CFD) models capture transient micro-turbulence, such approaches are computationally prohibitive for real-time shipboard control. In contrast, this study emphasizes topological feasibility or viability. A viable water-transfer path $p$ is defined as a sequence of vertices $p = (v_1, v_2, \ldots, v_k)$ that satisfies a set of operational constraints $R=R_{edge} \cup R_{path}$, namely, $R$ includes both edge-level and path-level feasibility rules. Edge-level constraints are related to whether a directed edge can be added to the graph, while path-level constraints determine whether an enumerated route can be accepted as a viable path.

More generally, edge-level constraints could be affected by pump direction, valve availability and known operational restrictions, while path-level constraints could include: (1) Boundary Constraint: An effective path sequence must originate from a source tank ($v_1 \in V_{tank}$) or a starting pump and terminate at a distinct destination tank ($v_k \in V_{tank}$). (2) Actuation Constraint: The path must intersect at least one active pump ($p \cap V_{pump} \neq \emptyset$) to provide the necessary water supply. By applying these constraints, the graph representation effectively filters out physically infeasible valve permutations. Each validated, effective path uniquely corresponds to a synchronized array of binary valve-operation commands. These commands are then encoded as a discrete action within the RL framework discussed in Section IV. Ultimately, an optimally trained RL agent learns to avoid ineffective paths, selecting instead the most efficient sequence from all feasible paths to complete the ballast operation.

To automate the generation of all feasible water-transfer paths, **Algorithm 1** first applies the edge-level constraint set $R_{edge}$ to construct the directed adjacency matrix; path-level constraints are then used to filter the enumerated candidate paths. After candidate paths are enumerated from the adjacency matrix, $R_{path}$ is applied to determine whether each path satisfies the boundary, actuation, and other known constraints. **Algorithm 1** is designed to transform any piping topology into a directed graph, from which all feasible water-transfer paths are generated. Using Fig.1 as an illustrative example, BP_01 and BP_02 are assumed to be the starting pumps for simplicity. By **Algorithm 1**, the ballast piping network can be automatically converted to all feasible paths (see Fig. 2). If there is any path violating this rule, it would be removed by the path-level constraints. After generating all the feasible paths, we would represent each path with a directed graph, which is then stored as computational data structures such as an adjacency matrix or adjacency list.

The topological abstraction can be understood via the green curve highlighted in Fig. 1, which mirrors the pathway in the topmost row of Fig. 2(a) starting from pump BP_02, routing through valves SW_709V and BW_05, and ending at target tank DB_07. Across the system architecture, path depths fluctuate, commonly encompassing four, five, or more nodes. By translating this physical layout into a directed graph, the ballast tracking task is reformulated from an intractable valve-combination challenge

**Algorithm 1:** Pseudocode for automatically generating all possible water-transfer paths from ballast water piping topology

**Input:** Component list $C$, Pipe segment list $L$, Operational constraints $R = R_{edge} \cup R_{path}$

**Output:** Directed graph $G = (V, E)$, Adjacency matrix/list Adj

**Initialization:** $V \leftarrow \emptyset$; $E \leftarrow \emptyset$; Adj ← empty dictionary

**for** each component $c \in C$ do

  Create node $v_c$

  Store node attributes: ID, component type, location, operational status

  $V \leftarrow V \cup \{v_c\}$; $Adj[v_c] \leftarrow \emptyset$

**end for**

**for** each pipe segment $l \in L$ do

  Extract the connected components $c_{start}$ and $c_{end}$

  $V \leftarrow V \cup \{v_c\}$; $Adj[v_c] \leftarrow \emptyset$

**end for**

**for** each pipe segment $l \in L$ do

  Extract the connected components $c_{start}$ and $c_{end}$

  Decide flow directions according to: pump direction, valve orientation, source-to-tank transfer direction, operational rules

  **if** flow is allowed from $c_{start}$ to $c_{end}$ then

    $E \leftarrow E \cup \{(c_{start}, C_{end})\}$

    $Adj[c_{start}] \leftarrow Adj[c_{start}] \cup \{c_{start}\}$

  **end if**

  **if** reverse flow is allowed from $c_{end}$ to $c_{start}$ then

    $E \leftarrow E \cup \{(c_{end}, c_{start})\}$

    $Adj[c_{end}] \leftarrow Adj[c_{end}] \cup \{c_{start}\}$

  **end if**

**end for**

Apply edge-level operational constraints $R_{edge}$

Remove edges that violate pump direction, valve availability, or known operational rules

Apply path-level operational constraints $R_{path}$

Remove the path that violates boundary constraints and actuation constraints

Return $G = (V, E)$, $Adj$

into a discrete path-selection problem, optimizing the action space for the RL agent.

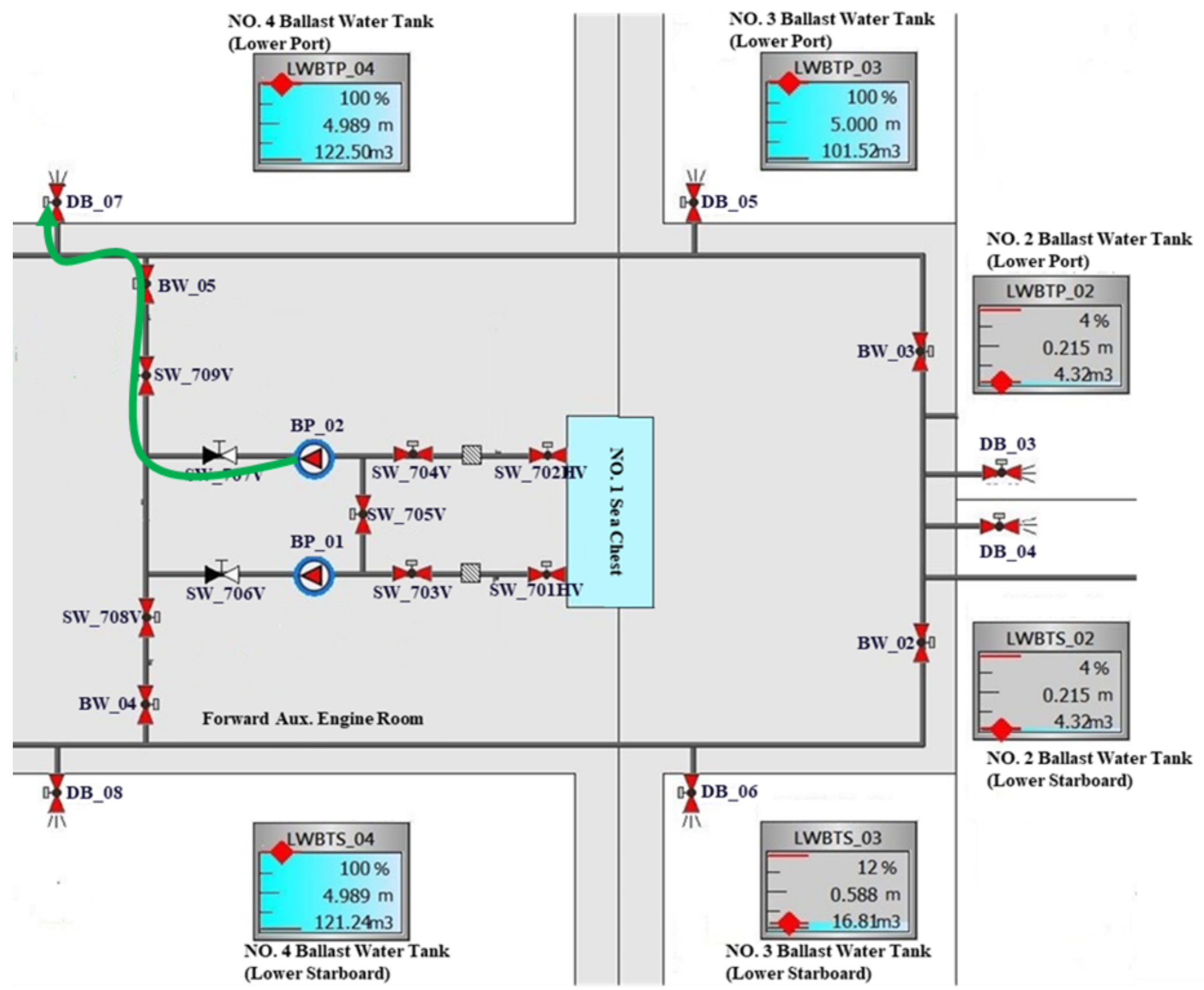

Figure 1. An exemplar blast piping layout. The green curve represents the path at the topmost row in Fig.2(a).

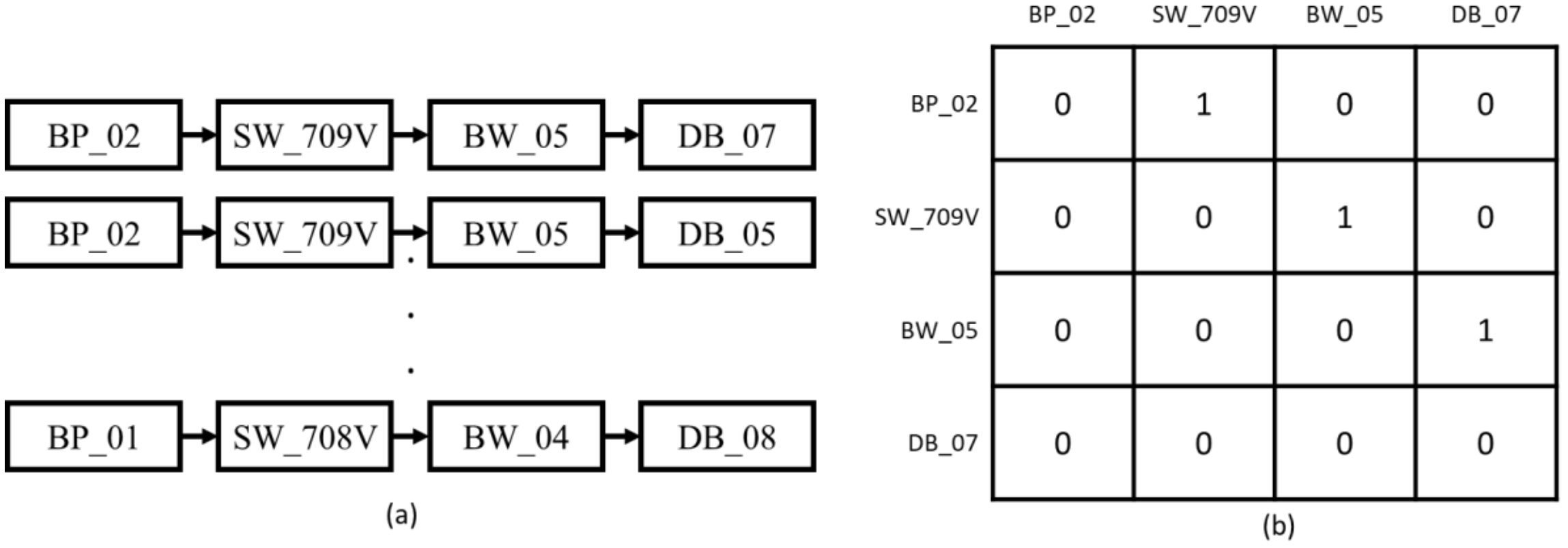

Figure 2. (a) The directed graph of all feasible paths generated by applying **Algorithm 1** to the piping topology in Fig.1; (b) The adjacency matrix corresponding to the feasible path at the top row of Fig.2(a).

Topological viability does not guarantee successful ballast transfer under unmodeled blockage conditions. A path may satisfy all predefined topological and operational constraints but still produce no-flow behavior due to localized blockage, sediment accumulation, or valve malfunction. If all hydraulic and blockage states were directly observable, the system could be modeled as an MDP. However, because blockage locations and latent valve conditions cannot be directly monitored from tank-level measurements, the practical decision problem is inherently partially observable. Solving the exact POMDP would require maintaining and updating a belief state over all possible hidden blockage configurations, which is computationally prohibitive for the proposed ballast-routing setting. Therefore, this study adopts a practical MDP approximation by augmenting observable tank-level states with recent action-outcome histories. This approximation allows the RL agent to select feasible transfer paths based on available observations while preserving computational tractability for the control loop, effectively augmenting raw tank data with a temporal sequence of past action outcomes. This design also enables the agent to assess system variables at each interval, select an optimal discrete path, and learn from subsequent tank-level differentials and reward feedback. The following section elaborates the MDP formulation, covering states, actions, reward engineering, transitions, and terminal conditions.

### *B. Approximating POMDP as MDP*

A standard POMDP solver requires maintaining and continuously updating a belief state, which represents a probability distribution over all possible hidden parameters. For a ballast system with $N$ valves, the combinatorial space of hidden blockage states reaches $2^N$ combinations, even under single-blockage assumptions, and scales exponentially in multi-blockage scenarios. Calculating and updating this massive belief state matrix at each decision step far exceeds the computational limits of onboard hardware, introducing unacceptable latency bottlenecks for critical stability operations. Consequently, directly solving a high-dimensional POMDP is computationally prohibitive, making it unsuitable for real-time inference on resource-constrained shipboard edge computing devices.

To overcome this bottleneck, we draw inspiration from the frame stacking strategy in DQN, where the issue of partial observability is tackled by augmenting tank-level observations with recent action-outcome histories. Thus, we can cast the autonomous fluid transfer problem as an MDP, denoted by the tuple ($S$, $A$, $R$, $P$, $\gamma$). By defining the state space strictly around observable macroscopic hydraulic metrics, we maintain computational tractability for the control loop, while explicitly delegating the resolution of hidden topological states (i.e., blockage locations) to the diagnostic module detailed in Section IV. The specific components of our MDP are defined as follows:

**State Space (S):** To ensure vessel safety, the observation state $s_t \in S$ at time step $t$ must reflect both the internal hydraulic conditions and the external vessel attitude. We define the continuous state vector as $\boldsymbol{s_t = [l_1, l_2, \ldots, l_m]^T}$, where $l_i \in (0,100\%]$ represents the normalized liquid level of the $i$-th ballast tank. However, because internal pipe blockages are unobservable latent variables, a single static snapshot provides insufficient information to detect stagnant flow. To restore a pseudo-Markovian property, we apply a frame-stacking mechanism. The augmented state vector $s_t \in S$ fed into the RL agent is constructed by concatenating the current observation with the history of preceding $k$ steps of observations and actions. Formally, it is expressed as: $\boldsymbol{s_t = [o_t, a_{t-1}, o_{t-1}, a_{t-2}, o_{t-2}, \ldots, a_{t-k}, o_{t-k}]}$. In this study, the temporal window length is set to $k = 4$. This sequential trajectory provides critical hydraulic transition gradients, allowing the agent to implicitly infer flow rates and path blockages. Note that global vessel stability (heel and trim) is managed at the supervisory level by defining symmetric target states (e.g., filling Port and Starboard tanks simultaneously), allowing the RL agent to focus exclusively on optimal hydraulic routing and fault avoidance.

**Action Space (*A*):** To achieve computational tractability and ensure physical validity, we employ a Depth-First Search (DFS) algorithm (Tarjan, 1972) based on the adjacency matrix to pre-calculate all feasible paths. An action $a_t \in A$ thus corresponds to selecting a specific valid path, which translates into a synchronized array of valve actuations, drastically reducing the dimensionality of the action space.

**Reward Function (*R*):** The reward function is the core driver of the agent's routing efficiency and diagnostic function. Instead of a simple scalar, the immediate reward $r_t$ is engineered as a multi-objective piecewise function to balance task completion, mechanical preservation, and fault identification. Denote $f$ as the number of target tanks are full in the current step; $n$ the number of targeted tanks; $s$ the number of tanks with a static water level, and $i$ the number of target tanks successfully be filled in current step. Therefore, the number of intended recipient tanks is given by $n = f + i + s$. The switching indicator is defined as $t = 0$ while the current action is same as last action, otherwise it would be 1.

$$r_t = \begin{cases} R_{succ}, & \text{if all target tanks are full} \\ R_{fill}, & \text{if } f > 0 \\ P_{clog}, & \text{if } n = s \\ \alpha i - \beta s - \lambda t, & \text{otherwise} \end{cases}, \quad (1)$$

where $R_{succ}$ and $R_{fill}$ represent the reward if all target tanks are already full when f>0, respectively; $P_{clog}$ is activated only when the intended tanks experience stagnation; Eq. (1) is designed to provide parametric insensitivity while preserving the intended preference order among different operating outcomes. First, $R_{succ}$ must be greater than the intermediate filling reward $R_{fill}$, considering the completion of all target ballast tanks represents the ultimate objective of the ballast task. Second, the no-flow penalty $P_{clog}$ should have the most negative reward value in Eq. (1). In other words, any action that results in a measurable filling progress should receive a higher reward than a completely ineffective clogged condition, even if the action incurs static-tank or switching penalties. In short, Eq. (1) must satisfy $R_{succ} > R_{fill}$ and $P_{clog} < \alpha i - \beta s - \lambda t < R_{succ}$. If these two constraints are not met, the agent may learn an undesirable policy and avoid changing actions to reduce the switching penalty, even when no target ballast tank receives water. By assigning the clogged no-flow condition the largest negative penalty, the reward function encourages the agent to abandon ineffective routes and search for alternative feasible transfer paths. Unless otherwise stated, the default setting for ($R_{succ}$, $P_{clog}$, $\alpha$, $\beta$, $\lambda$) is (10, -0.5, 0.01, 0.07, 0.05).

### *C. Optimization Objective and Operational Assumptions*

To formulate a computationally tractable model while preserving the core hydrodynamic behaviors necessary for training the RL agent, the following operational assumptions are established for a simulated maritime environment: (1) **Incompressible Fluid and Quasi-Steady Flow**: The ballast water is treated as an incompressible fluid. The hydrodynamic transitions between states are approximated as quasi-steady, focusing on macroscopic volumetric changes and static stability rather than transient micro-turbulence, (2) **Binary Valve Actuation**: For topological pathfinding, the controllable valves ($V_{cont}$) are modeled as binary mechanisms (i.e., fully opened or fully closed). Partial throttling is beyond the scope of this discrete action space and is reserved for future continuous-control studies; (3) **Sensor Noise Abstraction**: While real-world tank level sensors are susceptible to wave-induced sloshing, this preliminary framework abstracts external environmental noise to isolate and validate the topological blockage prediction mechanism. Building upon this MDP framework, the ultimate objective of the reinforcement learning agent is to discover an optimal policy $\pi^*$ that maps states to actions in a manner that maximizes the expected cumulative discounted reward. Mathematically, this optimization objective is defined as:

$$J(\pi) = E_\pi\left[\sum_{t=0}^{\infty} \gamma^t r_t\right], \tag{2}$$

where $\gamma \in [0,1]$ is the discount factor that dictates the agent's foresight, prioritizing immediate stability and task completion while ensuring long-term sequence validity. Eq. (2) serves as the foundation for training DQN detailed in Section IV.

## IV. METHODOLOGY

To solve the topology-constrained MDP defined in Section III and maximize the expected return $J(\pi)$ of Eq. (2), traditional heuristic controllers, such as Proportional-Integral-Derivative (PID) loops or PLC-based hardcoded logic, are inadequate. This insufficiency stems from their inability to adapt to the highly non-linear topology transformations induced by stochastic pipe blockages. To address this limitation, we introduce **RL-Ballast**, a framework combining reinforcement learning (RL) with other custom modules (green blocks in Fig.3). While this work utilizes a DQN as the exemplar RL agent for illustrative purposes, the framework architecture is agnostic to the underlying RL algorithm and can seamlessly accommodate alternative models. As illustrated in Fig. 3, the RL-Ballast framework architecturally comprises six interconnected modules:

1. **Environment Module**: Models the ballast piping topology, generates feasible fluid-transfer paths to establish the discrete action space, simulates tank-level state transitions, and injects stochastic blockage conditions during training and evaluation. To stabilize and accelerate convergence, a performance-based blockage curriculum is integrated into this module, which dynamically scales the difficulty of the blockage scenarios based on the agent's moving task success rate.
2. **Frame-Stacking Module**: Augments the instantaneous system observation with a temporal sequence of historical states. Because internal pipe blockages are latent variables not directly observable via localized tank-level measurements, a static state snapshot provides insufficient information to differentiate viable transfer paths from obstructed ones. By temporally stacking of consecutive observations and action-outcome pairs, the agent can infer the underlying hydrodynamic state and assess whether previous actions yielded meaningful volumetric progress.
3. **DQN-based Agent** (the red block in Fig.3): Receives the stacked state representation to estimate the state-action value function (Q-value) across all valid transfer-path actions. This agent adapts the canonical DQN architecture by tailoring two core components: Experience Replay and Target Networks. During the training phase, the agent balances exploration and exploitation via an $\varepsilon$-greedy policy, updating its parameterized policy network through experience replay sampling and periodic target network synchronization. During deterministic inference, the policy selects the action maximizing the estimated Q-value within the constrained action space.
4. **Episodic Failure Memory**: Records actions executed and flagged as ineffective within the current episode. Specifically, if a selected transfer path fails to generate a measurable variance in tank levels, the corresponding action is logged as a failure. This module captures transient, fault-related information localized to the current deployment episode without necessitating parameter updates or fine-tuning of the primary DQN weights.
5. **Action Masking Module**: Leverages the data logged in the Episodic Failure Memory to dynamically prune the action space, preventing the agent from re-selecting previously obstructed paths. Consequently, actions associated with fully processed tanks or verified transfer failures are excluded from the candidate action set during inference.
6. **Diagnostic Module**: Converts failed transfer histories into blockage-candidate rankings. When a transfer path produces no measurable tank-level increase, the module records the involved valves and pipe segments and updates their failure counts. Components that repeatedly appear in failed paths receive higher suspicion scores.

Together, these modules allow RL-Ballast to perform adaptive routing while simultaneously accumulating physically interpretable evidence for blockage-candidate diagnosis.

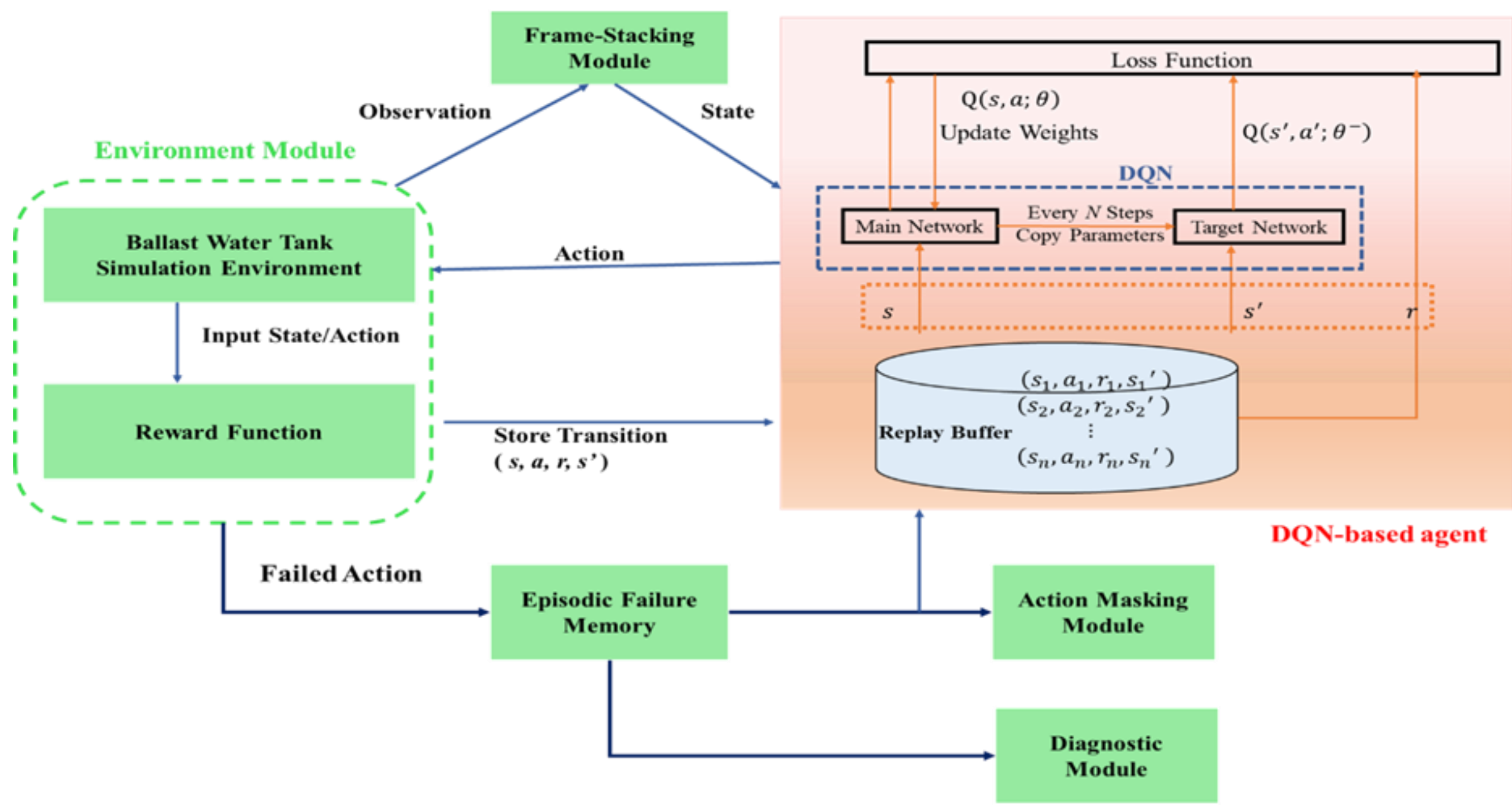

Figure 1. Integrating DQN with frame-stacking module, environment module, episodic failure memory, action masking, and diagnostic module to construct **RL-Ballast.**

*A. Action Space Pruning and System Workflow of RL-Ballast*

The performance of **RL-Ballast** depends on how well the DQN can approximate the optimal action-value function $Q^*(s,a) \approx Q(s,a;\theta)$, where $\theta$ represents the weights of the neural network. Recall that the computational challenge in applying RL to complex piping systems is the combinatorial explosion of the action space. Thus, feeding all possible valve permutations into the DQN would severely hinder network convergence. To address this bottleneck, the output layer of our DQN avoids issuing individual valve commands altogether. Instead, it leverages the DFS algorithm, operating on the system's adjacency matrices, to pre-calculate a discrete set of valid topological paths $A_{valid}$. Consequently, the DQN only needs to output Q-values for these dynamically feasible paths, reducing the action space dimensionality by several orders of magnitude.

As indicated in Fig. 3, the DQN is designed to be strictly governed by the physical parameters of the ballast system. The complete training process, which ensures that the DQN conforms to the ballast-water planning constraints, is detailed in **Algorithm 2**. To capture non-linear hydraulic relationships and state transitions, the network utilizes two fully connected hidden layers with Rectified Linear Unit (ReLU) activations. Crucially, to resolve the combinatorial explosion of raw valve permutations, the output layer dimension is explicitly constrained to match the 54 feasible paths predefined by the DFS algorithm. A linear activation function is applied at this terminal layer to generate continuous Q-value estimates for each discrete topological action, thereby

**Algorithm 2:** Pseudocode of DQN-based agent Training for Ballast Water Path Planning

**Input:** Ballast piping graph $G$, action set $A$, maximum episodes $E$, maximum steps $T$
**Initialize** DQN network $Q(s,a;\ \theta)$
**Initialize** target network $Q(s,a;\theta^-)$
**Initialize** replay buffer $D$
**for** episode = 1 to $E$ do
  Reset environment and obtain initial state $s_0$
    **for** $t = 1$ to $T$ do
      Select action at using $\epsilon$-greedy policy
      Execute action at in the ballast environment
      Observe reward $r_t$, next state $s_{t+1}$ and terminal flag done
      Store transition $(s_t, a_t, r_t, s_{t+1}, done)$ in $D$
      Sample a minibatch from $D$
      Update $\theta$ by minimizing the temporal-difference loss (TD error)
      Periodically update target network $\theta^-$
      **if** done **then**
        **break**
      **end if**
    **end for**
**end for**

ensuring computational efficiency for shipboard edge deployment. The network is trained by minimizing the loss function $L(\theta)$ using a custom experience replay buffer $D$:

$$L(\theta) = \mathbb{E}_{(s,a,r,s')\sim D}\left[\left(r + \gamma \max_{a'} Q(s,a;\theta^-) - Q(s,a;\theta)\right)^2\right], \tag{3}$$

where $\theta^-$ denotes the parameters of a periodically updated target network, ensuring training stability in the simulated environment. Note that the standard DQN is known to suffer from an overestimation bias due to the use of Eq. (3) that invokes the same target network to both select and evaluate an action. To mitigate this issue and further enhance learning stability, we also implemented a DDQN variant as a comparative baseline. DDQN decouples action selection from evaluation by utilizing a main network $(\theta)$ to select the greedy action, and a target network ($\theta^-$) to evaluate its value.

$$Y^{DDQN} = r + \gamma Q(s', argmax_{a'} Q(s',a';\ \theta^-);\ \theta^-) \tag{4}$$

In Eq. (4), the loss function for the DDQN agent replaces the standard target with $Y^{DDQN}$. Both DQN and DDQN architectures are evaluated in our experiments to demonstrate the implementation flexibility of RL-Ballast. Direct optimization of the policy $\pi^*$ over a high-dimensional state space to maximize the expected cumulative return $J(\pi)$ is computationally intractable. Our framework circumvents this bottleneck through an indirect optimization pathway. We systematically minimize a surrogate loss function, $L(\theta)$ to reduce the Bellman error, which drives the neural network's Q-value estimations toward the true optimal action-value function, $Q^*$. Once convergence is achieved, executing a greedy policy with respect to these optimized Q-values naturally yields the optimal policy, $\pi^*$ thereby satisfying the global maximization objective $J(\pi)$.

Fig. 4 shows the complete system workflow of RL-Ballast. Unlike traditional deterministic controllers, RL-Ballast operates through a fundamentally decoupled scheme comprising an online interaction phase and an offline optimization phase. In the offline optimization phase, the framework strictly exploits the off-policy nature of the DQN model. Rather than updating weights synchronously after every action step, a randomized minibatch of historical transitions is periodically sampled from the replay buffer. The weight network parameters $\theta$ are iteratively updated via backpropagation to minimize the temporal difference (TD) against a periodically updated target network ($\theta^-$). Such decoupling stabilizes the learning convergence against the high variance of hydrodynamic interactions, while effectively isolating the intensive model optimization workload from real-time operational constraints.

Completely decoupled from this background learning process, the online interaction phase governs real-time execution. The system first perceives the environment by capturing the current hydrodynamic observation to construct the state vector $s_t$. Then, the DQN agent evaluates Q-values exclusively for the topologically valid flow paths pre-calculated via DFS. To balance environmental exploration with exploitative routing, an $\varepsilon$-greedy policy dictates the selected action $a_t$ which is then translated into physical valve actuations. As the simulator executes this configuration, the environment transitions to a subsequent state $s_{t+1}$ and generates an immediate reward $r_t$ governed by the multi-objective hydraulic criteria. This complete interaction is then compressed into a transition tuple ($s_t$, $a_t$, $r_t$, $s_{t+1}$) and securely appended to the Experience Replay buffer. As a result, the framework in a sense executes an implicit fault diagnosis: if the emitted reward matches the exact stagnation penalty $P_{clog}$, the diagnostic module (shown in Fig.4 the composed of Failed-Action Memory and Scoring Blockage Suspicion) immediately flags the executed path to update the blockage suspicion scores. Advantageously, this diagnostic routine operates as an independent branch, entirely bypassing the neural network's gradient flow to preserve real-time computational efficiency.

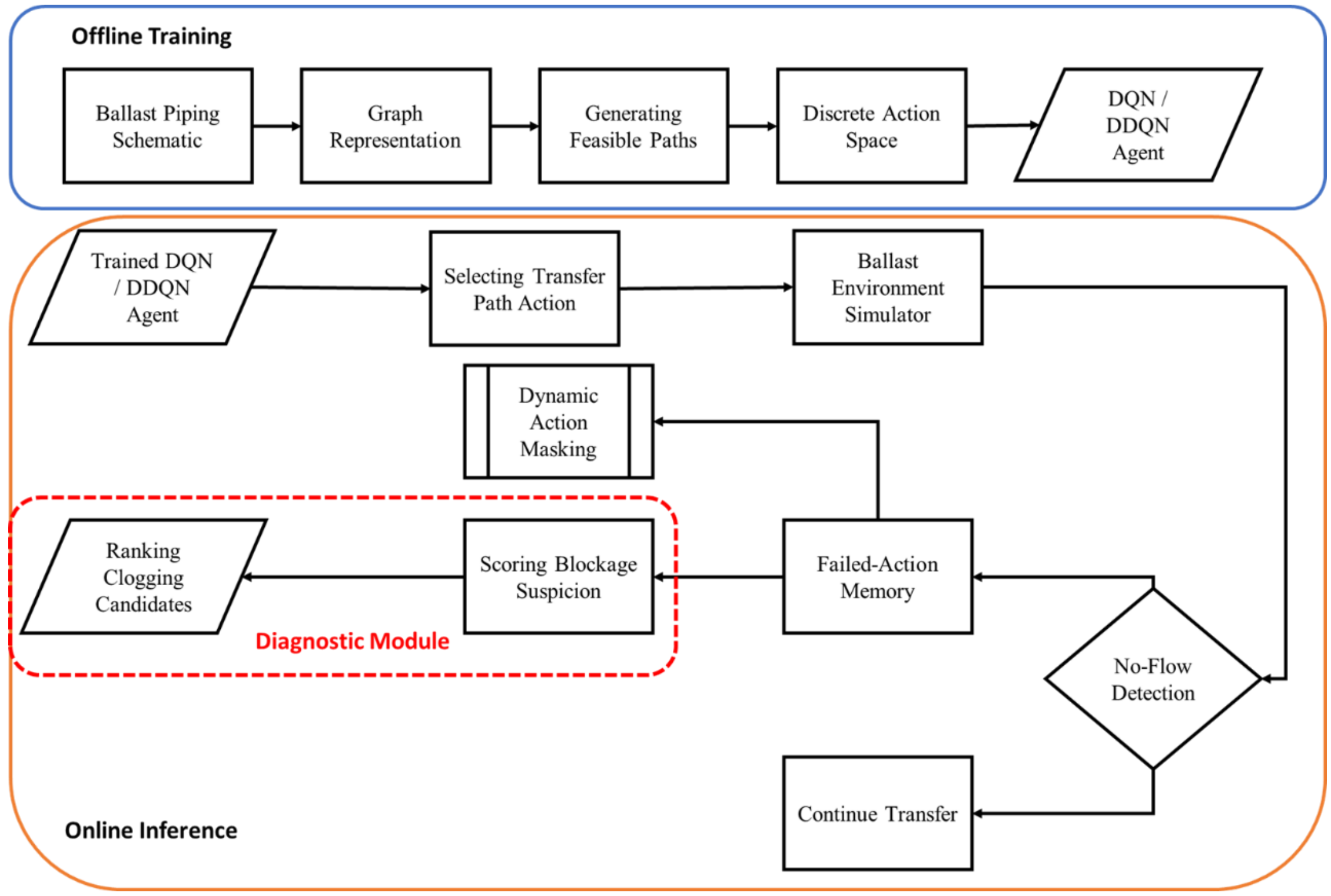

Figure 4. The complete workflow of RL-Ballast based on DQN-based agent during the training phase and inference phase. .

### *B. Performance-Based Blockage Curriculum Training*

Training a reinforcement learning agent directly under highly complex blockage conditions may lead to unstable learning, excessive exploration failures, and slow convergence. To address this issue, this study adopts a performance-based blockage curriculum training strategy (Bengio et al., 2009; Portelas et al., 2020). Instead of exposing the agent to the maximum number of blockage components from the beginning, the training environment gradually increases the blockage difficulty according to the recent learning performance of the agent. In our implementation, we adopt a straightforward approach to achieve performance-based curriculum learning. If the moving success rate exceeds a predefined threshold and the standard deviation is lower than a predefined threshold, the number of blocked components is increased. At each training episode, a set of blocked components is randomly sampled from the candidate valves or pipe segments. The number of blocked components is controlled by the current blockage level. During the early training stage, the blockage level is kept low so that the agent can first learn basic feasible path selection and water-transfer behavior. After the agent maintains a high task success rate over a fixed evaluation window, the blockage level is increased by one until the predefined maximum blockage level is reached.

### *C. Failure History-Based Blockage Candidate Scoring*

Although the RL agent can learn to avoid ineffective transfer paths, operators still require interpretable diagnostic information to understand why a selected path fails. To address this need, this study proposes a failure history-based blockage candidate scoring method. Instead of attempting to directly localize the blockage from dense pressure or flow measurements, the proposed method uses the accumulated failed transfer paths to rank candidate valves or pipe segments that are repeatedly associated with no-flow outcomes. The diagnostic computation is lightweight because it relies on frequency counting over the components contained in failed paths rather than high-fidelity fluid dynamics simulation or additional neural-network inference. When a selected transfer path produces no measurable tank-level increase and receives the no-flow penalty, the components along that path are recorded as suspicious candidates. Components that appear more frequently in failed paths receive higher blockage suspicion scores. Therefore, the diagnostic module provides a ranked list of likely blockage candidates that can support operator inspection and maintenance prioritization under limited sensing conditions. This scoring process operates independently from the DQN weight-update process. As a result, the failure history-based diagnostic module can provide interpretable blockage evidence without interfering with the RL agent's policy optimization or online action selection. The resulting output should be interpreted as a relative suspicion ranking rather than an exact blockage localization result.

### *D. Dynamic Action Masking for Inference Deadlock Avoidance via Negative Episodic Memory*

During the deterministic inference phase (where the exploration rate $\epsilon = 0$), DRL agents are highly susceptible to infinite loops, or inference deadlocks, when encountering unmodeled topological blockages. This vulnerability stems from the state aliasing under partial observability: if a selected flow path is obstructed, the macroscopic physical state (i.e., tank levels) remains unchanged ($s_{t+1}=s_t$). Without the stochasticity of exploration, the static Q-network repeatedly evaluates the same failing action as optimal, fundamentally trapping the control loop. To overcome this critical limitation, we draw theoretical inspiration from the paradigm of Model-Free Episodic Control (MFEC) introduced by Blundell *et al.* (2016). While traditional MFEC accelerates learning by caching high-reward experiences for one-shot adaptation, we invert this principle to establish a Negative Episodic Memory mechanism via dynamic action masking. Specifically, when an executed action $a_t$ triggers the definitive stagnation penalty ($P_{clog}$), the control architecture instantaneously caches this critical failure and prunes $a_t$ from the valid action space $A_{valid}$. Consequently, in the subsequent decision step, the agent is forced to select the topologically feasible action with the next highest Q-value. This algorithmic intervention functions as an instantaneous episodic constraint, seamlessly breaking state-aliasing deadlocks. Ultimately, it ensures continuous, fail-safe ballast operations without incurring the computational overhead of complex state-space augmentation, rendering it highly viable for real-time shipboard edge computing. The complete inference procedure is summarized in **Algorithm 3.** At each decision step, the trained DQN first evaluates the remaining candidate actions after excluding the actions stored in the failed-action memory. If the selected action leads to a no-flow response, the action is added to the failed-action memory and the components along the failed path are counted for blockage candidate scoring. This procedure allows the agent to reroute within the same episode while simultaneously accumulating diagnostic evidence for ranking suspicious blockage candidates. The blockage candidate scoring equation is given as follows:

$$Score(c) = \sum_{p_i \in F} 1(c \in p_i), \tag{5}$$

where $F$ denotes as the union of failed paths, $c$ denotes as valve or pipe segment candidate.

**Algorithm 3:** Pseudocode of Inference with Failed-Action Memory and Dynamic Action Masking

```
Input: Trained DQN Q (s, a; θ), action set A, maximum steps T
Initialize failed-action memory M = ∅
Initialize component failure counter C(v) = 0 for all components
Reset environment and obtain initial state s_0
for t = 1 to T do
    Select action a_t = arg max_{a∈A\M} Q(s_t, a; θ)
    Execute action a_t
    Observe tank-level response
    if no-flow condition is detected then
        Add a_t to failed-action memory M
        for each component v in path (a_t) do
            C(v) = C(v)+1
        end for
    end if
    Update state s_{t+1}
    if target tanks are filled then
        break
    end if
end for
Compute blockage suspicion score for each component
Return selected action history and ranked blockage candidates
```

## V. Experiments and results

This section systematically evaluates the effectiveness of RL-Ballast through a series of simulation experiments. All experiments were conducted using the TensorFlow framework on a workstation equipped with an Intel Core i9-9900K CPU and an NVIDIA RTX 2080 GPU. The software environment included CUDA 10.0, cuDNN v7.6.5, and Python 3.6.9.

### *A. Experiment Setup*

RL-Ballast was evaluated in a simulated ballast-water environment comprising 6 tanks, 12 valves, 2 pumps, and 54 feasible transfer paths generated from the graph. Unless otherwise stated, in both training and evaluation, the number of target tanks was set to three, corresponding to half of the six ballast tanks. This setting provides a representative transfer task while avoiding overly trivial single-tank filling cases. Fig. 5 shows a simulated scenario having the ballast-water tanks and piping routes, as well as the selected blockage candidates, including SW_708V, SW_709V, BW_02, BW_03, BW_04, and BW_05. Based on engineering knowledge of the ballast piping layout, several safety-critical components in Fig. 5 are excluded from the blockage-candidate set. This does not imply that these components cannot fail, but rather that such failures would represent severe system-level faults requiring inspection or maintenance instead of online rerouting. Since two simultaneous blockages would disable one-third of the six predefined blockage candidates, such a condition is treated as severe degradation beyond ordinary rerouting. Therefore, the maximum number of simultaneous blockages is limited to one as a controlled preliminary setting, while more complex multi-blockage scenarios are reserved for future work. Unknown blockage scenarios were simulated by disabling selected valves or pipe segments during evaluation. Evaluation metrics include task success rate, average steps, average reward per episode, repeated failed actions, and blockage-candidate ranking hits. The experiments were designed to answer the following questions: (1) Can a DQN-based agent complete ballast transfer under normal conditions? (2) Can the agent reroute water transfer under unknown blockage conditions? (3) Is a DQN-based agent faster than rule-based algorithms? (4) Does dynamic action masking reduce repeated ineffective actions? (5) Can failed transfer histories identify likely blockage candidates?

Table 1 lists the hyperparameters for training RL-Ballast. The maximum number of decision steps in each episode was set to 200. If the agent completed the target ballast transfer task before reaching this limit, the episode was regarded as successful. Otherwise, the episode was terminated when the maximum step limit was reached. The agent was trained for 150,000 episodes to allow sufficient interaction with different routing and blockage configurations. The replay buffer size was set to 100,000 transitions, and a mini-batch size of 32 was used for each training update. The learning rate was set to 0.0001, and the discount factor $\gamma$ was set to 0.98, allowing the agent to consider both immediate transfer progress and long-term task completion. The target network was updated every 3,000 training steps to stabilize Q-value estimation. An $\varepsilon$-greedy exploration strategy was used during training. The initial exploration rate was set to 0.9 to encourage broad exploration of feasible transfer paths at the beginning of training. The exploration rate was gradually reduced to 0.01, allowing the agent to shift from exploration to exploitation as training progressed. During inference, $\varepsilon$ was set to 0 so that the trained policy selected actions deterministically according to the learned Q-values and the applied action-masking rules. This setting ensures that the evaluation results reflect the learned routing policy rather than random exploratory behavior.

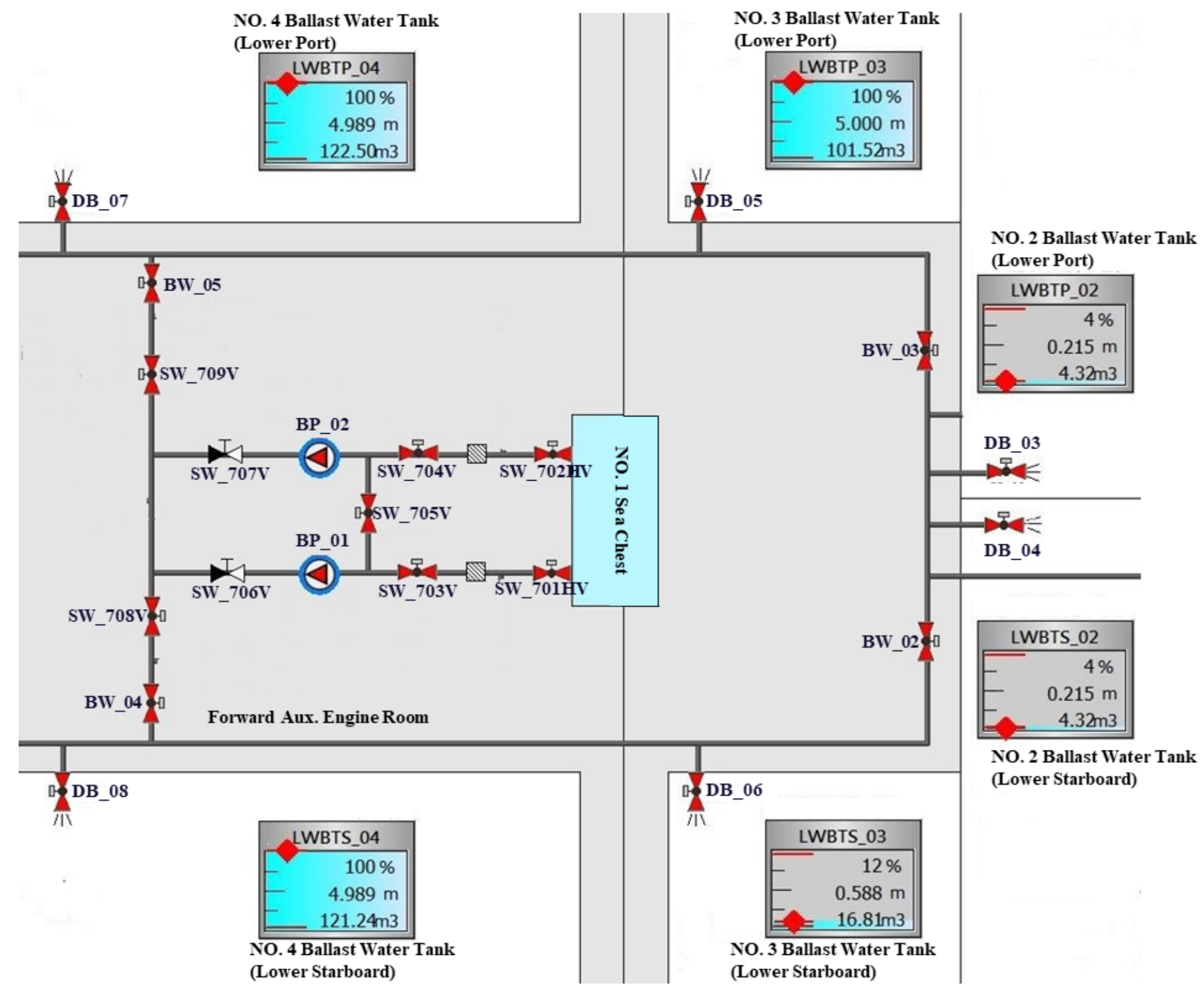

Figure 5. A simulated scenario of ballast water tanks and piping routes, where possible blockage points are SW_708V, SW_709V, BW_02, BW_03, BW_04 and BW_05.

**Table 1** Hyperparameters for training RL-Ballast

| Parameter | Value | Parameter | Value |
|---|---|---|---|
| Number of ballast tanks | 6 | Learning rate | 0.0001 |
| Number of valves | 12 | Discount factor $\gamma$ | 0.98 |
| Number of feasible actions | 54 | Initial exploration rate $\epsilon$ | 0.9 |
| Maximum steps per episode | 200 | Final exploration rate $\epsilon$ | 0.01 |
| Training episodes | 150000 | Inference exploration rate $\epsilon$ | 0 |
| Replay buffer size | 100000 | Target network update interval | 3000 |
| Batch size | 32 | Frame-stacking length | 4 |

*B. Training Convergence Analysis*

To evaluate whether the proposed DQN-based agent can learn an effective ballast-water path-selection policy, the training convergence behavior was analyzed using episode total reward, average reward, task success rate, and average steps. The agent was trained with different random seeds, and the average results were reported. Among these metrics, task success rate, average reward, and average steps were used as the primary indicators to examine both learning stability and ballast-transfer performance. Fig. 6(a) shows the task success rate during training. The success rate declined between 10,000 and 20,000 episodes due to the performance-based curriculum learning strategy. When the number of blockages increased from 0 to 1, the agent needed to explore a new routing policy. However, the success rate quickly recovered to nearly 1, indicating that the DQN-based agent learned an effective policy under the no-blockage scenario and could adapt to blockage conditions.

The reward function defined in Eq. (1) significantly influences the learning behavior of the RL agent. In this study, the reward is designed to distinguish successful ballast transfer from ineffective transfer attempts without hard-coding a specific routing solution. To evaluate the robustness of the reward design, two reward-parameter settings are evaluated to examine whether Eq. (1)

can consistently converge under different coefficient settings. As mentioned earlier, the first setting is the default ($R_{succ}$, $P_{clog}$, $\alpha$, $\beta$, $\lambda$) = (10, -0.5, 0.01, 0.07, 0.05). The second setting ($R_{succ}$, $P_{clog}$, $\alpha$, $\beta$, $\lambda$) = (30, -1, 0.015, 0.08, 0.05).

As shown in Fig. 6(b), the total reward per episode is presented using Min–Max-normalized moving-average curves. For visualization, each moving-average reward curve is independently normalized to [0, 1] as Eq. (6).

$$\bar{R}_{norm}(e) = \frac{\bar{R}(e) - \min_e \bar{R}(e)}{\max_e \bar{R}(e) - \min_e \bar{R}(e)}, \tag{6}$$

where $\bar{R}(e)$ denotes the moving-average reward at the $e$-th episode. This normalization is useful in examining convergence trends across different reward-parameter settings. In the early training stage, the normalized reward remains relatively low because the agent frequently selects inefficient or ineffective transfer paths. As training progresses, the moving-average reward gradually increases, indicating that the agent learns to select actions that lead to successful tank-level increases while avoiding repeated ineffective routes. Despite that the two settings use different in coefficient values, they all comply with the same constraints defined in Eq. (1). As a result, the variations in coefficient values only affect the convergence speed and reward magnitude, the learning behaviors of these two settings largely follow the same pattern.

*C. Baseline Routing without Blockage*

The first experiment evaluates the basic path-planning capability of the trained agents under normal operating conditions. In this setting, all valves, pumps, and pipe segments are functional, and no blockage is introduced into the ballast piping network. The purpose is to verify whether DQN and DDQN can learn feasible ballast-transfer sequences and complete the target tank-level task before evaluating more challenging blockage scenarios. For comparison, a deterministic Dijkstra shortest-path strategy (Dijkstra, 1959) was implemented as the rule-based baseline. Since the ballast-water system is modeled as a directed graph, this baseline provides a reproducible benchmark for selecting minimum-cost transfer paths based on fixed edge weights. However, it cannot learn from past outcomes or adapt its routing policy through interaction. Success rate, average steps, and average reward are used as evaluation metrics. Success rate measures task completion, average steps evaluate decision efficiency, and average reward reflects the learned policy quality of the RL agents. Since the rule-based baseline does not optimize a learned reward function, its average reward is reported as N/A. As shown in Table 2, all three methods achieve a 100% success rate under the no-blockage condition. However, DQN and DDQN complete the task in 41 and 38 average steps, respectively, compared with 60 steps for the rule-based baseline. DDQN also achieves a higher average reward than DQN, increasing from 22.6 to 24.9. These results indicate that the graph-based action representation enables the RL agents to learn valid and efficient path-selection policies under nominal ballast operating conditions.

**Table 2** Baseline Routing Performance without Blockage

| Method | Success Rate (%) | Avg. Steps | Avg. Reward |
|---|---|---|---|
| DQN | 100% | 41 | 22.6 |
| DDQN | 100% | 38 | 24.9 |
| Rule-based | 100% | 60 | N/A |

*D. Comparison with Rule-Based Algorithms*

The second experiment evaluates whether the trained agents can complete the water-transfer task under unknown single-blockage scenarios and compares their performance with a deterministic rule-based baseline. The evaluation metrics include success rate, average steps, average reward, and wasted action rate. Success rate, average steps, and average reward are defined in the same manner as in the previous experiment, while the wasted action rate measures the proportion of ineffective actions. As shown in Table 3, all methods achieve a 100% success rate under single-blockage conditions. However, DQN and DDQN require fewer average steps than the rule-based baseline, completing the task in 42.167 and 41.5 steps, respectively, compared with 61 steps for the rule-based method. This indicates that the learned policies improve routing efficiency rather than merely finding feasible paths. Compared with DQN, DDQN further reduces the average steps from 42.167 to 41.5, increases the average reward from 19.493 to 24.1, and lowers the wasted action rate from 9.48% to 6.02%. Although the rule-based baseline has a lower wasted action rate of 1.64%, it requires more decision steps and cannot learn action preferences from previous interactions. These results show that DDQN provides a more efficient learned routing policy under unknown blockage conditions.

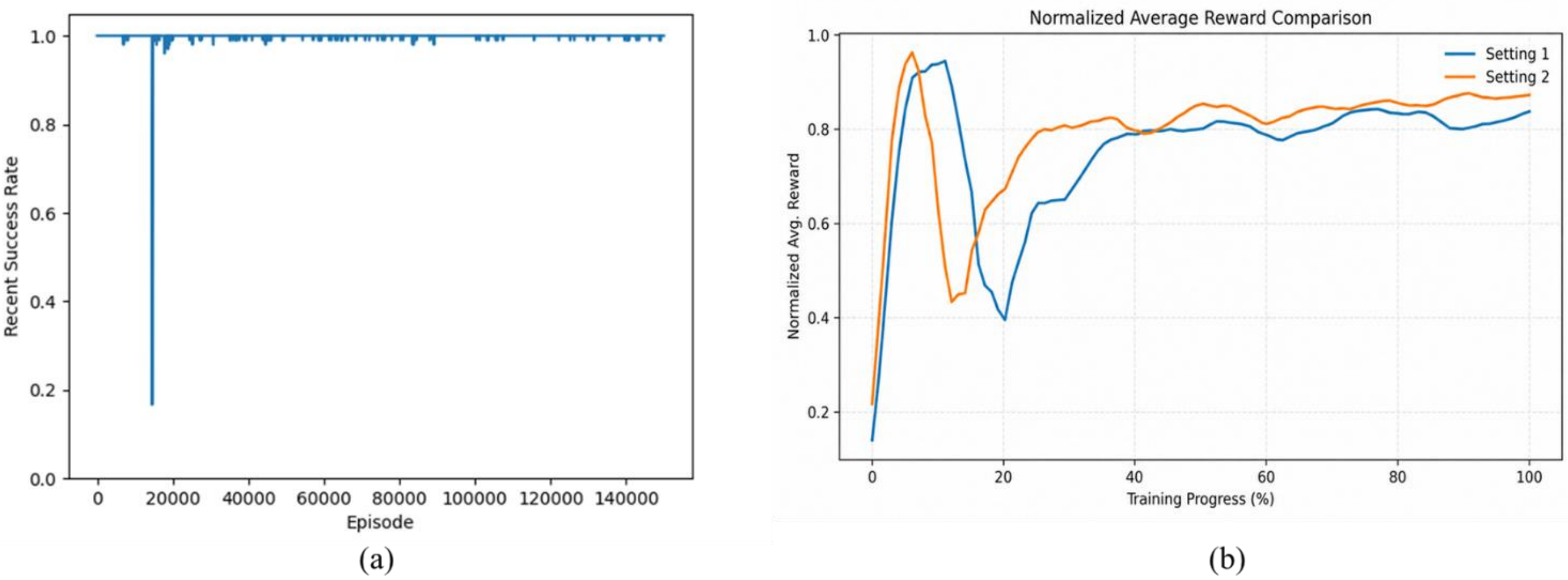

(a) (b)

Figure 6. (a) the success rate per episode; (b) the Min-Max normalized average reward per episode of two different parametric settings.

**Table 3** Baseline Performance with Single-Blockage

| Method | Success Rate (%) | Avg. Steps | Avg. Reward | Wasted Action Rate (%) |
|---|---|---|---|---|
| DQN | 100% | 42.167 | 19.493 | 9.48% |
| DDQN | 100% | 41.5 | 24.1 | 6.02% |
| Rule-based | 100% | 61 | N/A | 1.64% |

*E. Dynamic Rerouting under Unmodeled Blockages*

Next, we evaluate the agent's ability to adapt to unknown blockage conditions. Here, DDQN is used because it achieved better overall performance. Selected valves or pipe segments were blocked without revealing their locations to the agent. The agent could only infer blockage from tank-level responses after each action. The number of target tanks was fixed at three, and the target tanks were randomly selected in each trial. Each blockage scenario was tested 1,000 times, and the average metrics were reported. In addition to the previously defined metrics, average failed actions are used to measure how many ineffective transfer attempts occur before the agent identifies and avoids obstructed routes. As shown in Table 4, the DDQN agent maintains a 100% success rate across all six single-blockage scenarios. The average number of failed actions is 2.5, the average number of steps is 41.5, and the average reward is 0.5815. Among the tested blockage components, BW_02 results in the highest number of failed actions, with 6 failed actions, 43 average steps, and an average reward of 0.5279. In contrast, SW_708V and BW_04 require only 1 failed action and 40 average steps, achieving the highest average reward of 0.6175. These results indicate that RL-Ballast can infer blockage effects from tank-level responses and autonomously reroute ballast-water transfer under unknown blockage conditions.

**Table 4** Rerouting Performance under Unknown Blockage

| Blockage Component | Success Rate (%) | Avg. Failed Actions | Avg. Steps | Avg. Reward |
|---|---|---|---|---|
| SW_708V | 100% | 1 | 40 | 0.6175 |
| SW_709V | 100% | 2 | 42 | 0.5738 |
| BW_02 | 100% | 6 | 43 | 0.5279 |
| BW_03 | 100% | 3 | 42 | 0.5786 |
| BW_04 | 100% | 1 | 40 | 0.6175 |
| BW_05 | 100% | 2 | 42 | 0.5738 |
| **Average** | **100%** | **2.5** | **41.5** | **0.5815** |

*F. Ablation Study of Dynamic Action Masking*

To evaluate the contribution of dynamic action masking, an ablation study was conducted by comparing four DDQN methods: no masking, tank-completion masking, failed-action masking, and the proposed combined masking strategy. The purpose is to determine whether excluding previously failed actions improves fault-aware inference under unknown blockage conditions. In this experiment, the number of target tanks was kept at the default setting, while both target tanks and blockage points were randomly selected. Each method was tested under every blockage candidate, and the average results are reported in Table 5. In addition to the previously defined metrics, repeated failed action counts measure how often the agent reselects ineffective actions, while diagnostic feasibility qualitatively evaluates whether the resulting failure histories are useful for blockage-candidate diagnosis. Using only tank-completion masking results in a lower success rate of 83.33%, 27.33 repeated failed actions, 67.67 average steps, and a high wasted action rate of 42.11%, because the agent can still select paths containing blocked components. Using only failed-action masking achieves a 100% success rate and zero repeated failed actions, but increases the average steps to 94.83 and the wasted action rate to 37.78%. In contrast, the proposed combined masking strategy achieves a 100% success rate, zero repeated failed actions, 41.5 average steps, and a wasted action rate of 3.61%, while maintaining high diagnostic feasibility. Although DDQN without masking achieves fewer average steps of 38.67 and a lower wasted action rate of 1.29%, its diagnostic feasibility is only medium because the generated failure histories may be less interpretable. Overall, the combined masking strategy provides the best balance between task completion, routing efficiency, and physically interpretable diagnostic records.

**Table 5** Ablation Study of Dynamic Action Masking

| Method | Success Rate (%) | Average Repeated Failed Action Counts | Avg. Steps | Wasted Action Rate (%) | Diagnostic Feasibility |
|---|---|---|---|---|---|
| DDQN without masking | 100% | 0 | 38.67 | 1.29% | Medium |
| DDQN + tank-completion mask | 83.33% | 27.33 | 67.67 | 42.11% | Low |
| DDQN + failed-action mask | 100% | 0 | 94.83 | 37.78% | High |
| Proposed method | 100% | 0 | 41.5 | 3.61% | High |

*G. Blockage Candidate Ranking*

The final experiment evaluates whether failed transfer histories can provide useful diagnostic information. Here, the trained DDQN agent with the combined action masking strategy was used. Different blockage points were defined as different scenarios; for example, S1 simulates a blockage at SW_708V. For each scenario, the proposed scoring method ranks valves or pipe segments according to their accumulated failure counts. The diagnostic performance is evaluated using Top-k hits, where a prediction is a hit if the true blockage is one of the Top-k ranked candidates

$$Top - k_{hit} = \frac{N^{(k)}_{\mathrm{correct}}}{N_{\mathrm{total}}} \times 100\%, \tag{7}$$

where $N^{(k)}_{\mathrm{correct}}$ denotes the number of blockage scenarios in which the true blocked component is one of the top-(k) ranked candidates, and $N_{\mathrm{total}}$ denotes the total number of evaluated blockage scenarios. When multiple components obtain the same failure score, they are treated as a tied suspicious group, and the diagnostic result should be interpreted as a candidate ranking rather than a unique fault localization result. As shown in Table 6, the proposed failure history-based scoring method achieves a Top-$3_{(hit)}$ rate of 100% across all evaluated blockage scenarios. The Top-$1_{(hit)}$ rate is (4/6=66.7%), because the true blockage is ranked first in four out of six scenarios. For serially connected components, such as SW_708V/BW_04 and SW_709V/BW_05, the method can identify the correct suspicious blockage group but may not always distinguish the exact blocked component. This is because upstream and downstream blockages on the same transfer branch can produce indistinguishable tank-level responses. To account for this ambiguity, we also report a tie-adjusted top-1 hit rate (Top-$1_{(tie\text{-}hit)}$). When multiple components obtain the same highest suspicion score, they are treated as a tied suspicious group. If the true blockage is uniquely ranked first, the Top-1 credit is 1; if it is included in a tied group (G), a fractional credit of (1/|G|) is assigned; otherwise, the credit is 0. Under this definition, the tie-adjusted Top-$1_{(hit)}$ rate is 83.3%, because the two serially connected blockage cases receive a fractional credit of 0.5 instead of 0. Therefore, the diagnostic output should be interpreted as a ranked suspicious region or candidate group rather than exact component-level localization in all cases.

**Table 6** Blockage Candidate Ranking Performance

| **Scenario** | **True Blockage** | **Top-1(hit)** | **Top-1 (tie-hit)** | **Top-3(hit)** | **Rank of True Blockage** |
|---|---|---|---|---|---|
| S1 | SW_708V | 1.0 | 1.0 | 1.0 | 1 |
| S2 | SW_709V | 1.0 | 1.0 | 1.0 | 1 |
| S3 | BW_02 | 1.0 | 1.0 | 1.0 | 1 |
| S4 | BW_03 | 1.0 | 1.0 | 1.0 | 1 |
| S5 | BW_04 | 0 | 0.5 | 1.0 | 2 |
| S6 | BW_05 | 0 | 0.5 | 1.0 | 2 |
| | **Average** | **0.67** | **0.833** | **1.0** | **N/A** |

## VI. DISCUSSIONS

From a practical perspective, RL-Ballast provides a decision-support mechanism for fault-aware ballast-water operation under limited sensing conditions. Under the broader Shipping 4.0 paradigm, intelligent onboard systems are expected to improve not only navigation autonomy but also the reliability and efficiency of internal shipboard operations. This direction is also consistent with the broader sustainability motivation of SDG 13 and climate-change mitigation, where digitalization and AI-enabled decision support can contribute to safer, more efficient, and less wasteful maritime operations. In this context, RL-Ballast focuses on the internal ballast-water transfer process by reducing repeated ineffective route selection and supporting blockage-aware operational decisions. The modular design of RL-Ballast provides implementation flexibility. Although DQN and DDQN are adopted in this study as representative value-based agents, the framework itself is not restricted to a specific reinforcement learning algorithm. The environment module, graph-based action generator, state representation, episodic failure memory, action masking mechanism, and blockage candidate scoring module can be coupled with alternative RL architectures when the control objective or action space changes. In the present formulation, the DQN/DDQN is suitable because the graph-based representation converts the raw valve-combination problem into a finite set of discrete transfer-path actions.

The primary limitation of the current framework lies in the spatial resolution of blockage localization. Because the diagnostic module relies mainly on macroscopic tank-level responses and failed path histories, it may not always distinguish the exact blocked component when multiple valves or pipe segments are serially connected and repeatedly appear in the same failed transfer paths. In such cases, the proposed method can identify a suspicious component group or pipe region, but the exact component-level blockage location may still require additional sensing, inspection, or active diagnostic testing. Therefore, the diagnostic output should be interpreted as a relative blockage suspicion ranking rather than an exact localization result or calibrated fault probability.

Future work will exploit the algorithmic flexibility of RL-Ballast to extend the framework toward more realistic and complex ballast-control scenarios. Continuous-action reinforcement learning may be explored for precise valve throttling or variable pump-speed regulation, while hierarchical or multi-agent reinforcement learning may be considered for larger ballast networks. Physics-informed or constrained reinforcement learning may also be introduced to incorporate hydraulic feasibility, topological invariants, and operational safety constraints into the learning process. Finally, additional pressure, flow, vibration, or acoustic sensing may be integrated with failed-action-based diagnosis to improve the localization accuracy for serially connected blockage candidates.

## VII. CONCLUSION

This paper proposed RL-Ballast, a graph-based reinforcement learning framework for adaptive path planning and blockage candidate scoring in ship ballast-water systems. The study was motivated by the need for intelligent onboard internal systems under the Shipping 4.0 paradigm, where autonomous and reduced-crew vessels require reliable control of safety-critical shipboard operations. By modeling the ballast piping system as a topology-constrained graph, feasible water-transfer routes were generated and converted into discrete reinforcement learning actions. This design transforms the original valve-control problem into a feasible path-selection problem, reducing invalid action combinations while preserving the physical constraints of the piping network. Based on this graph-generated action space, DQN and DDQN agents were trained to select ballast-water transfer paths using observable tank-level states and action-outcome histories. A performance-based blockage curriculum was introduced to improve training stability, while episode-level failed-action memory and dynamic action masking were used during inference to avoid repeated ineffective paths under unknown blockage conditions. In addition, failed transfer histories were reused as diagnostic evidence by accumulating failure counts over shared valves and pipe segments. The resulting diagnostic output provides a ranked list of suspicious blockage candidates rather than a calibrated fault probability or exact component-level localization.

Simulation results show that RL-Ballast can complete all evaluated unexpected single-blockage scenarios with a 100% success rate. Compared with the Dijkstra rule-based baseline, DDQN reduces the average number of decision steps from 61.0 to 41.5, indicating improved routing efficiency under blockage conditions. The ablation study further shows that the proposed combined masking strategy achieves a 100% success rate, zero repeated failed actions, 41.5 average steps, and a wasted action rate of 3.61%, providing a better balance between task completion, routing efficiency, and diagnostic interpretability. For blockage candidate scoring, the proposed failure history-based method achieves a 100% top-3 hit rate, a 66.7% strict top-1 hit rate, and an 83.3% tie-adjusted top-1 hit rate under serially indistinguishable blockage conditions. These results suggest that reinforcement learning can support not only adaptive ballast-water rerouting but also maintenance-oriented diagnostic assistance under limited sensing conditions. Since the present study is based on a simulated single-blockage setting, future work will extend the framework toward higher-fidelity hydraulic simulation, multi-blockage scenarios, sensor-noise modeling, and hardware-in-the-loop validation.

**CRediT Author Statement**

Ming-Kuan Lin: Conceptualization, Methodology, Software, Validation, Formal analysis, Investigation, Visualization, Writing – original draft. Yi-Chung Lai: Methodology, Software, Validation, Formal analysis, Writing – review & editing. Ming-Hsin Chiang: Methodology, Software, Investigation. Jung-Hua Wang: Conceptualization, Methodology, Supervision, Project administration, Writing – review & editing. Tsung-Wei Pan: Writing – review & editing.

**Data Availability**

Simulation data and pertaining source code are available upon the acceptance of the paper and reasonable request.

## REFERENCE

Ai, Z., Cao, H., Wang, M., Yang, K., 2024. Ship Ballast Water System Fault Diagnosis Method Based on Multi-Feature Fusion Graph Convolution. *J. Phys. Conf. Ser. 2755*, 012028. https://doi.org/10.1088/1742-6596/2755/1/012028

Bengio, Y., Louradour, J., Collobert, R., Weston, J., 2009. Curriculum learning, in: *Proceedings of the 26th Annual International Conference on Machine Learning*, ICML '09. Association for Computing Machinery, New York, NY, USA, pp. 41–48. https://doi.org/10.1145/1553374.1553380

Blundell, C., Uria, B., Pritzel, A., Li, Y., Ruderman, A., Leibo, J.Z., Rae, J., Wierstra, D., Hassabis, D., 2016. Model-Free Episodic Control. https://doi.org/10.48550/arXiv.1606.04460

Dijkstra, E. W. (1959). A note on two problems in connexion with graphs. *Numerische Mathematik, 1*, 269–271. https://doi.org/10.1007/BF01386390

Ferreno-Gonzalez, S., Diaz-Casas, V., Miguez-Gonzalez, M., San-Gabino, C.G., 2025. Detection of Pipe Ruptures in Shipboard Firefighting Systems Using Machine Learning and Deep Learning Techniques. *Appl. Sci. 15*, 1181. https://doi.org/10.3390/app15031181

Geng, Y., Liu, E., Wang, R., Liu, Y., 2020. Deep Reinforcement Learning Based Dynamic Route Planning for Minimizing Travel Time. https://doi.org/10.48550/arXiv.2011.01771

Hasselt, H. van, Guez, A., Silver, D., 2016. Deep Reinforcement Learning with Double Q-Learning. *Proc. AAAI Conf. Artif. Intell. 30*. https://doi.org/10.1609/aaai.v30i1.10295

Hu, C., Cai, J., Zeng, D., Yan, X., Gong, W., Wang, L., 2020. Deep reinforcement learning based valve scheduling for pollution isolation in water distribution network. *Math. Biosci. Eng. 17*, 105–121. https://doi.org/10.3934/mbe.2020006

Kalashnikov, D., Irpan, A., Pastor, P., Ibarz, J., Herzog, A., Jang, E., Quillen, D., Holly, E., Kalakrishnan, M., Vanhoucke, V., and Levine, S., "Scalable Deep Reinforcement Learning for Vision-Based Robotic Manipulation," *Proceedings of the 2nd Conference on Robot Learning, PMLR*, vol. 87, pp. 651–673, 2018.

Kufoalor, D.K.M., Wilthil, E., Hagen, I.B., Brekke, E.F., Johansen, T.A., 2019. Autonomous COLREGs-Compliant Decision Making using Maritime Radar Tracking and Model Predictive Control, in: *2019 18th European Control Conference (ECC)*. pp. 2536–2542. https://doi.org/10.23919/ECC.2019.8796273

Li, Y., Wu, Defeng, You, Z., Chen, G., Wu, Dongjie, 2025. Deep reinforcement learning for collision avoidance in unmanned surface vehicles: State-of-the-art. *Appl. Ocean Res. 164*, 104778. https://doi.org/10.1016/j.apor.2025.104778

Portelas, R., Colas, C., Weng, L., Hofmann, K., Oudeyer, P.-Y., 2020. Automatic Curriculum Learning For Deep RL: A Short Survey [WWW Document]. arXiv.org. URL https://arxiv.org/abs/2003.04664v2 (accessed 5.30.26).

Sarhadi, P., Naeem, W., Athanasopoulos, N., 2022. A survey of recent machine learning solutions for ship collision avoidance and mission planning: *IFAC Conference on Control Applications in Marine Systems, Robotics and Vehicles. Proc. 14th IFAC Conf. Control Appl. Mar. Syst. Robot. Veh. CAMS 2022*, IFAC-PapersOnLine 257–268. https://doi.org/10.1016/j.ifacol.2022.10.440

Sullivan, B.P., Arias Nava, E., Desai, S., Sole, J., Rossi, M., Ramundo, L., Terzi, S., 2021. Defining Maritime 4.0: Reconciling principles, elements and characteristics to support maritime vessel digitalisation. *IET Collab. Intell. Manuf. 3*, 23–36. https://doi.org/10.1049/cim2.12012

Sullivan, B.P., Desai, S., Sole, J., Rossi, M., Ramundo, L., Terzi, S., 2020. Maritime 4.0 – Opportunities in Digitalization and Advanced Manufacturing for Vessel Development. *Procedia Manuf. 42*, 246–253. https://doi.org/10.1016/j.promfg.2020.02.078

Sutton, R.S., Barto, A.G., 2018. *Reinforcement Learning: An Introduction*, 2nd ed, Adaptive Computation and Machine Learning series. MIT Press, Cambridge, MA, USA.

Terven, J., 2025. Deep Reinforcement Learning: A Chronological Overview and Methods. *AI 6*, 46. https://doi.org/10.3390/ai6030046

Velasco-Gallego, C., Lazakis, I., Cubo-Mateo, N., 2024. Development of a Hierarchical Clustering Method for Anomaly Identification and Labelling of Marine Machinery Data. *J. Mar. Sci. Eng. 12*, 1792. https://doi.org/10.3390/jmse12101792

Wang, W., Cao, X., Gonzalez-Garcia, A., Yin, L., Hagemann, N., Qiao, Y., Ratti, C., Rus, D., 2023. Deep Reinforcement Learning Based Tracking Control of an Autonomous Surface Vessel in Natural Waters. https://doi.org/10.48550/arXiv.2302.08100